\pdfoutput=1
\documentclass[11pt]{article}

\usepackage[final]{acl}

\usepackage{times}
\usepackage{latexsym}

\usepackage[T1]{fontenc}
\usepackage[utf8]{inputenc}

\usepackage{microtype}

\usepackage{inconsolata}

\usepackage{graphicx}
\definecolor{ForestGreen}{RGB}{34,139,34}

\usepackage{amsmath}
\usepackage{booktabs}
\usepackage{tcolorbox}
\usepackage{listings}
\usepackage{array}
\usepackage{enumitem}
\usepackage{arydshln}

\lstdefinelanguage{JavaScript}{
  keywords={function, return, var, let, const, if, else, for, while, switch, break, case, default, try, catch, throw, new, this, console},
  sensitive=true,
  comment=[l]{//},
  comment=[s]{/*}{*/},
  morestring=[b]",
  morestring=[b]',
}

\title{Skill Discovery for Software Scripting Automation\\via Offline Simulations with LLMs}

\author{
 \textbf{Paiheng Xu\textsuperscript{1}}\thanks{Work done during internship at Adobe Research},
 \textbf{Gang Wu\textsuperscript{2}\thanks{Corresponding authors: Gang Wu (gawu@adobe.com) and Xiang Chen (xiangche@adobe.com)}},
 \textbf{Xiang Chen\textsuperscript{2}\footnotemark[2]},
 \textbf{Tong Yu\textsuperscript{2}},
 \textbf{Chang Xiao\textsuperscript{2}},
\\
 \textbf{Franck Dernoncourt\textsuperscript{2}},
 \textbf{Tianyi Zhou\textsuperscript{1}},
 \textbf{Wei Ai\textsuperscript{1}},
 \textbf{Viswanathan Swaminathan\textsuperscript{2}}
\\
\\
\textsuperscript{1}University of Maryland, College Park,
\textsuperscript{2}Adobe Research
}

\begin{document}
\maketitle
\begin{abstract}
Scripting interfaces enable users to automate tasks and customize software workflows, but creating scripts traditionally requires programming expertise and familiarity with specific APIs, posing barriers for many users. While Large Language Models (LLMs) can generate code from natural language queries, runtime code generation is severely limited due to unverified code, security risks, longer response times, and higher computational costs.
To bridge the gap, we propose an offline simulation framework to curate a software-specific skillset—a collection of verified scripts—by exploiting LLMs and publicly available scripting guides. 
Our framework comprises two components: (1) task creation, using top-down functionality guidance and bottom-up API synergy exploration to generate helpful tasks; and (2) skill generation with trials, refining and validating scripts based on execution feedback.
To efficiently navigate the extensive API landscape, we introduce a Graph Neural Network (GNN)-based link prediction model to capture API synergy, enabling the generation of skills involving underutilized APIs and expanding the skillset's diversity.
Experiments with Adobe Illustrator demonstrate that our framework significantly improves automation success rates, reduces response time, and saves runtime token costs compared to traditional runtime code generation. 
This is the first attempt to use software scripting interfaces as a testbed for LLM-based systems, highlighting the advantages of leveraging execution feedback in a controlled environment and offering valuable insights into aligning AI capabilities with user needs in specialized software domains.
\end{abstract}

\section{Introduction}

Scripting interfaces in software applications play a pivotal role in extending the capabilities of software beyond their standard functionalities. They enable users to automate repetitive tasks, customize workflows, and integrate applications with other systems \cite{ousterhout1998scripting}.
Prominent software like Adobe Illustrator and Adobe Photoshop support scripting through ExtendScript, which is Adobe's extended version of JavaScript tailored for their applications.~\footnote{Illustrator: \url{https://ai-scripting.docsforadobe.dev/}. Photoshop: \url{https://helpx.adobe.com/photoshop/using/scripting.html}.} 
Similarly, Microsoft Office applications provide scripting interfaces based on JavaScript, allowing users to automate tasks within Excel, Word, and other Office programs. \footnote{\url{https://learn.microsoft.com/en-us/office/dev/add-ins/reference/javascript-api-for-office}}
These scripting interfaces expose Application Programming Interfaces (APIs) that allow scripts to interact with the software's internal functions and data structures.

Traditionally, creating scripts using these interfaces requires programming expertise and familiarity with the specific APIs of the software, posing a barrier for many users. 
With the strong code generation capacity of Large Language Models (LLMs) \cite{chen2021evaluating,bubeck2023sparks}, some solutions generate code based on user query during runtime \cite{gandhi2023natural,zhao-etal-2024-nl2formula}. 
However, such runtime generation approaches have notable limitations: 
(1) the generated code is unverified when presented to the users, leading to low-quality code that may not align with users' intentions and can introduce security risks through unintended behaviors; 
(2) they impose considerable runtime burden, including increased response times and token generation costs, particularly for applications with a large user base.

In this work, we propose using offline simulation to curate a software-specific skillset -- a set of scripts that automate tasks within the software. 
Then they can be retrieved during runtime to solve user queries. We use publicly available scripting guides and LLMs' knowledge about the software to create the skillset. 
The offline simulation consists of two LLM-based components: (1) task creation, which generates useful tasks within the software, and (2) skill generation with trials, translating the generated tasks into skills with execution feedback from previous trials.
For task creation, we introduce two simulation strategies that use the software's functionality information (top-down) and API information (bottom-up) from the publicly available scripting guide.
To more efficiently explore the vast number of APIs supported in the software, we define synergistic API pairs as APIs that can work together in existing skills. 
We construct a synergistic API graph and train a link prediction model using a Graph Neural Network (GNN) to capture both the semantic and structural patterns of existing synergistic API pairs.
This enables the model to generalize to unseen API pairs and assess their compatibility.
The synergy of APIs is further used to prompt LLMs to generate tasks that better elicit the software's internal functions and data structures by using more long-tailed APIs. 

We conduct comprehensive experiments using Adobe Illustrator as a testbed to evaluate our approach. 
Our findings demonstrate that our offline simulation framework significantly improves success rates and efficiency of automation task compared to traditional runtime code generation methods.
Our main contributions are:

\begin{itemize}[nosep, leftmargin=*]
    \item We propose a novel offline simulation framework for curating a software-specific skillset, leveraging LLMs and publicly available scripting guides. Our framework employs two simulation strategies—top-down functional guidance and bottom-up API synergy exploration—to generate tasks and scripts that cover a wide range of software functionalities.
    \item We introduce a new setup that leverages the software's API information to explore the capacity of the software. We propose to use a GNN-based link prediction model to capture the synergy between APIs, which encourages generating skills involving underutilized or long-tailed APIs, thereby expanding the diversity and utility of the skillset.
    \item To the best of our knowledge, this is the first attempt to use software scripting interfaces as a testbed for LLM-based systems. This approach highlights the advantages of obtaining direct execution feedback in a controlled environment and offers valuable insights into aligning AI capabilities with user needs in specialized software domains.
\end{itemize}

\begin{figure*}[htbp]
    \centering
    \includegraphics[width=0.95\linewidth]{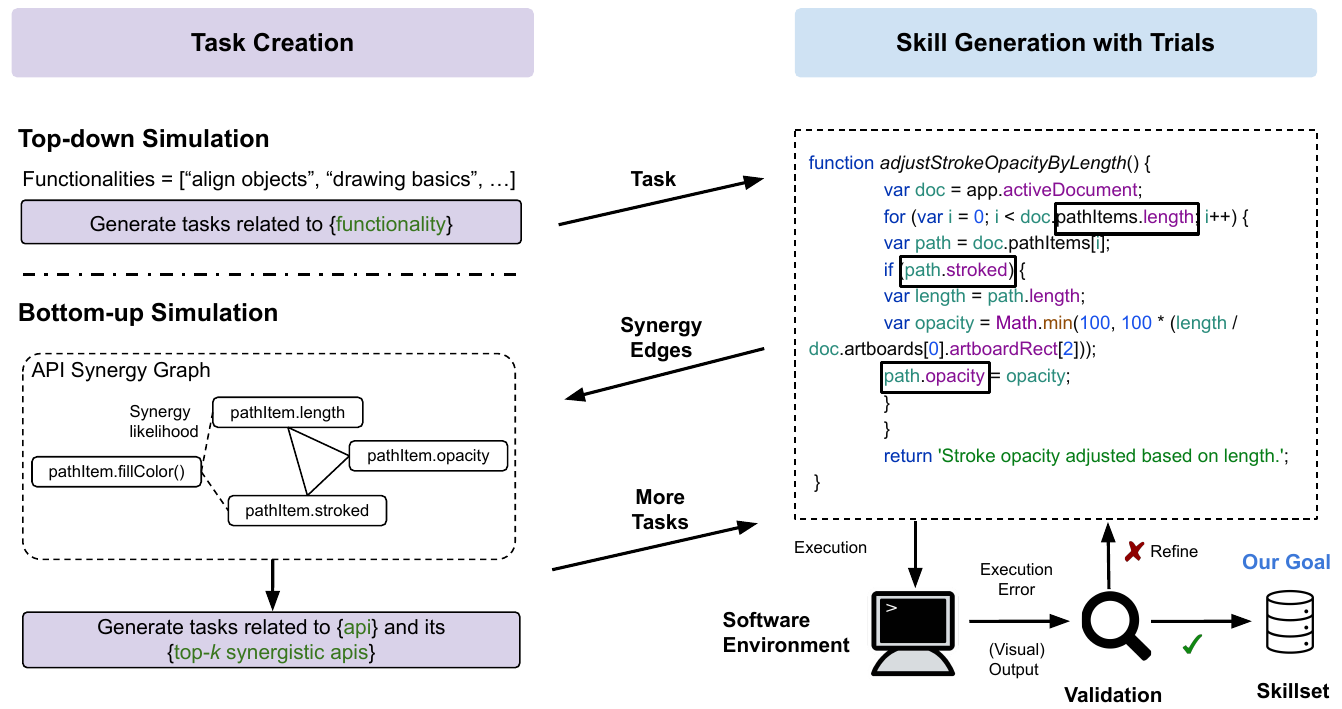}
    % \caption{Overall Structure \ph{briefly introduce the framework}}
    \caption{Overview of our offline simulation framework for skill discovery in software scripting automation. The framework consists of two components: (1) Task Creation (Section~\ref{sec:task_create}), utilizing two simulation strategies: top-down functionality guidance and bottom-up API synergy exploration to generate a wide range of tasks; and (2) Skill Generation with Trials (Section~\ref{sec:code_gen}), where LLMs iteratively refine scripts based on execution feedback to produce verified skills ready for runtime retrieval.}
    \label{fig:overview}
\end{figure*}
\section{Related Work}
\label{sec:related}
\subsection{Skill Discovery with LLMs}
LLM-based skill discovery is gaining attention across various domains, including embodied agents \cite{zhao2024agentic}, sandbox environment \cite{wang2024voyager}, and LLM tool usage \cite{qian2023creator,cai2024large,yuan2024craft,nguyen2024dynasaur}.
Similarly, skills in these scenarios are represented by code and designed to interact with corresponding environments.
However, existing methods either do not consider the coverage of these skills within their environments \cite{zhao2024agentic,wang2024voyager,qian2023creator,cai2024large} or rely on a rich dataset pairing task descriptions with desired outcomes in a question-answering format \cite{qian2023creator,cai2024large,yuan2024craft,nguyen2024dynasaur}.
Therefore, these approaches are limited in applicability to practical scenarios such as software scripting automation, where outcomes are more complex (sometimes involving multiple modalities) and cannot be easily encapsulated in a simple text string.
Curating a dataset with such diverse and complex outcomes is challenging, especially when the goal is to explore a software’s full automation potential. 
Our study addresses this gap by investigating novel skill discovery strategies that leverage a software’s publicly available functionalities and API information.

\subsection{Program Synthesis with LLMs}
Program synthesis aims to generate code given a natural language description \cite{zhang2023unifying}, 
where LLMs \cite{austin2021program,chen2021evaluating,nijkamp2023codegen} have demonstrated impressive performance.
Modern code generation models are typically evaluated on functional correctness \cite{liu2024your}, often requiring predefined unit tests \cite{chen2021evaluating, li2022competition}.
For example, pass$@k$ \cite{chen2021evaluating, li2022competition} evaluates the model’s chance of passing all unit tests with any of $k$ generated samples.
In the context of software scripting automation, execution results from the software environment provide immediate feedback on generated code.
Unlike the setting of code generation with given task descriptions, this study emphasizes deciding what tasks to generate in specific software, leveraging LLMs' strong code generation capacity that can be further improved with execution feedback from the software environment during offline simulation \cite{kim2024language, pan2024automatically}.

\subsection{Automation with LLM Agents}

With the rapid advancement of Large Language Models (LLMs), researchers have developed systems and benchmarks to automate complex tasks requiring multiple applications in computer environments \cite{xie2024osworld, cao2024spider2}. However, existing LLM-based agents face significant challenges in reliably automating these tasks, achieving only about a $15\%$ success rate across several hundred tasks.
Efforts to improve LLM capabilities have also focused on domain-specific tasks, such as spreadsheet manipulation \cite{li2024sheetcopilot, ma2024spreadsheetbench}, Graphical User Interface (GUI) automation \cite{gao2024assistgui, nguyen2025gui}, and web browsing \cite{yao2022webshop, ma2024agentboard, deng2024mind2web}. In these specialized domains, LLM-based agents demonstrate higher capacity.
Our paper aligns with this body of work by leveraging LLMs to automate domain-specific computer tasks. 
However, we present the first attempt to (1) automate tasks through the software's internal scripting environment, and (2) use LLM-based agents to systematically explore and identify which tasks can be automated in the software.
Our primary objective is to generate a verified skillset that represents a software's supported functionalities and aligns with users' practical needs, rather than develop LLM agents that directly automate tasks.
\section{Method}

% \begin{figure*}[htbp]
%     \centering
%     \includegraphics[width=0.95\linewidth]{latex/figures/skillgen_framework.pdf}
%     % \caption{Overall Structure \ph{briefly introduce the framework}}
%     \caption{Overview of our offline simulation framework for skill discovery in software scripting automation. The framework consists of two components: (1) Task Creation (Section~\ref{sec:task_create}), utilizing two simulation strategies: top-down functionality guidance and bottom-up API synergy exploration to generate a wide range of tasks; and (2) Skill Generation with Trials (Section~\ref{sec:code_gen}), where LLMs iteratively refine scripts based on execution feedback to produce verified skills ready for runtime retrieval.}
%     \label{fig:overview}
% \end{figure*}

In this section, we detail our framework for skill discovery with offline simulations, which can be divided into two modules, task creation and skill generation with trials.
We formulate the problem as follows: our goal is to develop a set of tasks (in natural language), $\mathcal{T} = \{t_1, t_2, ..., t_n\}$, which can be automated via the scripting interface in a software application. The corresponding scripts or code are defined as skills, $\mathcal{S} = \{s_1, s_2, ..., s_n\}$. 
After developing $S$ during offline simulation, a user can input some operation they want to automate as a query $q$ to the system during runtime, where a relevant skill, $s_i$, is retrieved to solve $q$. Figure \ref{fig:overview} shows an overview of the proposed framework and we now introduce our method in detail.

\subsection{Task Creation}
\label{sec:task_create}
We rely on LLMs to generate the tasks. To fully exploit the functionalities, we adopted two strategies to guide LLMs to ``search'' for possible tasks, i.e., a top-down approach that starts from high-level functionalities of the software, followed by a bottom-up approach that considers low-level APIs supported by the software.

\paragraph{Top-down}
We curate a list of high-level functionalities for the software. Taking Adobe Illustrator as an example, the high-level functionalities can be drawing, arranging objects, and so on. This list can be obtained from the content categorization from publicly available scripting guides. For each functionality, we prompt the LLMs to generate related tasks. Following previous works that adopt a long-term memory \cite{wang2024voyager,wang-etal-2024-llms-imaginarium}, we generate tasks in multiple rounds and integrate the feedback obtained from the previous round. We introduce what feedback our framework provides in Section \ref{sec:code_gen}. The prompts are in Appendix \ref{app:prompt}. We note that this step can also be considered as a warmup for the following bottom-up search.

\paragraph{Bottom-up}
A unique opportunity and challenge in the Software scripting interface setting is that we have a list of all APIs supported by the software, denoted as $\mathcal{A} = \{a_1, a_2, ..., a_n\}$, and the scripts largely rely on the ``collaboration'' of multiple APIs. Therefore, we prompt LLMs with appropriate API combinations to spark LLMs' knowledge about the software. 
Given the large number of APIs and the potential for multiple tasks to be accomplished using the same set of APIs, we introduce an API synergy graph, $\mathcal{G}$, to model the likelihood of APIs working together. 
Graph-based representations are widely employed to model relationships in related applications such as API recommendations \cite{qi2022correlation, huang20221+} LLM-based reasoning \cite{anokhin2024arigraph, liu2024explore}, and so on~\cite{wu2022graph, xu2025gfairhint}.

Specifically, we define APIs, $\mathcal{A}$, as the nodes in $\mathcal{G}$ and the edges, $\mathcal{E}$, represent whether the two APIs have appeared in a verified script. The node features are the semantic embeddings of the corresponding API descriptions, denoted as $\mathbf{X}$. The two nodes with a link are defined as a synergistic API pair that can work together. 
% \ph{We can introduce more complex molding later}. 
We then train a link prediction model by randomly masking existing links in $\mathcal{G}$ and predicting the likelihood of their existence. 
% We use a Graph Neural Network (GNN) model to achieve this. The model learns to predict the likelihood of a masked link by aggregating information from neighboring nodes and their features. Through training, the GNN captures both the semantic and structural patterns of synergistic API pairs, enabling it to generalize to unseen API pairs and assess their compatibility. \ph{This is a concise version}
We use a Graph Convolutional Network (GCN) model~\cite{kipf2017semi} to achieve this. The model aggregates information from neighboring nodes and their features, $\mathbf{H}^{(l)} = \text{GCN}^{(l)}(\mathbf{H}^{(l-1)}, \mathcal{E})$
where $\mathbf{H}^{(0)} = \mathbf{X}$ are the initial node embeddings and $\mathbf{H}^{(l)}$ are the node representations after $l$ layers of message passing.
The likelihood of an edge between two nodes $u$ and $v$ is modeled as:
$
\hat{y}_{uv} = \sigma\left(f\left(\mathbf{h}_u^{(L)}, \mathbf{h}_v^{(L)}\right)\right),
$
where $\mathbf{h}_u^{(L)}$ and $\mathbf{h}_v^{(L)}$ are the $L$-layer representations of nodes $u$ and $v$ (final layer), $f$ is a scoring function, such as the inner product $f(\mathbf{h}_u, \mathbf{h}_v) = \mathbf{h}_u^\top \mathbf{h}_v$, and $\sigma$ is the sigmoid function.
In short, the model learns to predict the likelihood of two API nodes working together by aggregating information from neighboring nodes and their features. 
The model is trained with a binary cross-entropy loss:
\begin{align*}
\mathcal{L} = -\frac{1}{|\mathcal{D}|} \sum_{(u, v) \in \mathcal{D}} & \Big[ y_{uv} \log(\hat{y}_{uv}) \\
& + (1 - y_{uv}) \log(1 - \hat{y}_{uv}) \Big],
\end{align*}
where $\mathcal{D}$ is the set of sampled edges (positive pairs from $\mathcal{E}$ and negative pairs not in $\mathcal{E}$), and $y_{uv} \in \{0, 1\}$ is the ground truth label for whether the edge $(u, v)$ exists.
After training, the GCN model captures both the semantic and structural patterns of synergistic API pairs, enabling it to generalize to unseen API pairs and assess their compatibility.

Then for each API $a_i$, 
% \ph{we can add some selections here}, 
we prompt LLMs to generate tasks related to $a_i$ and its top-$k$ synergistic APIs. We show full prompts in Appendix \ref{app:prompt}.

\subsection{Skill Generation with Trials}
\label{sec:code_gen}

% One advantage of generating skills offline is that we can generate the code with multiple trials offline without adding any burden to the users during runtime, providing a better user experience. 
% For each generated $t_i$, we use a strategy similar to the ideas of \citet{wang2024voyager, wang-etal-2024-llms-imaginarium}, where the LLM learns from the execution feedback in the software to refine its own outputs. 
% Additionally, we use another LLM or a Large Vision Language Model (LVLM) as a validator to judge the script by looking at the generated code, execution output, and (visual) outcome in the software.  
% The validator comments on whether the skill accomplishes the task and provides feedback for improvement for the next trial.
% A skill is added to the skillset $\mathcal{S}$ if it passes the validator. 

One advantage of generating skills offline is that we can generate the code with multiple trials offline without adding any burden to the users during runtime, providing a better user experience. 
For each generated $t_i$, we use a strategy similar to the ideas of \citet{wang2024voyager, wang-etal-2024-llms-imaginarium}, where the LLM learns from the execution feedback in the software to refine its own outputs. 
Additionally, we use another LLM or a Large Vision Language Model (LVLM) as a validator to judge the script by looking at the generated code, execution output, and (visual) outcome in the software.
The validator comments on whether the skill accomplishes the task and provides feedback for improvement for the next trial. 
We show the system and user prompt for the validator in Table \ref{tab:LVLM}.
A skill is added to the skillset $\mathcal{S}$ if it passes the validator.
Specifically, when refining the scripts, the LLM receives a structured prompt containing the task description, code from the previous round, any execution errors generated by the software, and the validator's assessment (including suggestions for improvement).
This structured feedback enables the LLM to refine the script by addressing both code-level issues and misalignments with task intent. We allow up to three trials per task.
We show the prompt for ExtendScript code generation in Table \ref{tab:trials}.
Example skills with above-mentioned elements are shown in Appendix \ref{app:example}.
\section{Experiments}
\label{sec:exp}

\paragraph{Adobe Illustrator as the Testbed} 
Adobe Illustrator is a leading vector graphics software used by professionals worldwide for tasks such as creating logos, illustrations, and complex design elements.
It supports multiple categories of high-level functionalities, including drawing, arranging objects, and applying effects. 
A full list of the high-level functionalities used in the top-down search in Section \ref{sec:task_create} is provided in Appendix \ref{app:func}, which can be obtained through Illustrator's official scripting guide. \footnote{\href{https://helpx.adobe.com/pdf/illustrator_reference.pdf}{https://helpx.adobe.com/pdf/illustrator\_reference.pdf}} 
Modern software platforms that support scripting interfaces offer tasks with varying levels of complexity.
This study focuses on generating ``atomic'' skills: fundamental, modular tasks that involve minimal design choices and serve as building blocks for more complex operations. 
For instance, in the context of Illustrator, instead of creating an entire flower design, an atomic skill would arrange pre-designed petals into a circular pattern, emphasizing precision and modularity.  
Illustrator's scripting interface supports $1818$ API endpoints, where $378$ are the methods and the remaining are attributes of the object, enabling extensive programmatic control over the software. 
We also explored using Excel as the testbed, but were unable to conduct a large-scale experiment due to limitations in Excel’s ability to permit programmatic control over Office Script from outside applications. See Appendix~\ref{app:excel} for details.
\begin{table}[t]
    \centering
    \footnotesize
    % \begin{tabular}{ >{\centering\arraybackslash}m{0.25\columnwidth} p{0.75\columnwidth}}
    \begin{tabular}{cc}
        \toprule
        Desc & arrange selected objects in circle  \\ \hline
        Code & \begin{minipage}[t]{0.7\linewidth}
        \vspace{-\baselineskip}
        \begin{lstlisting}[language=java, basicstyle=\ttfamily\tiny, backgroundcolor=\color{gray!10}, keywordstyle=\color{blue}, commentstyle=\color{green}, stringstyle=\color{red}]
function arrangeInCircle(cX, cY, radius) {
  var sel = app.selection
  if (sel.length === 0) {
    throw new Error('No selection');
    }
  var angleStep = 360 / sel.length;
  for (var i = 0; i < sel.length; i++) {
    var angle = angleStep*i*(Math.PI/180);
    var x = cX + radius * Math.cos(angle);
    var y = cY + radius * Math.sin(angle);
    sel[i].position = [x, y];
    }
  return 'Objects arranged in a circle';
}

        \end{lstlisting}
        \end{minipage}
        \\ \hline
        Effect & \includegraphics[width=0.1\linewidth]{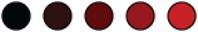} $\Rightarrow$ \includegraphics[width=0.2\linewidth]{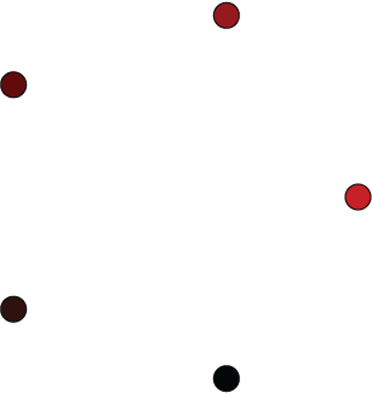} \\
        \bottomrule
        \hline
    \end{tabular}
    \caption{An example skill, \texttt{arrangeInCircle}. \texttt{Desc} is the natural language description. \texttt{Effect} shows the layout before and after running the skill.}
    \label{tab:example}
\end{table}

\paragraph{Simulation Setup}
Adobe Illustrator, as a vector design tool, focuses on creating and manipulating graphical objects.
In practical applications, a typical Illustrator project often contains numerous objects, and scripts are usually expected to operate on specific subsets of these objects. 
The common approach is to manually select the desired objects before running the script, leveraging the \texttt{selection} attribute available for each object.
Therefore, for each skill, we prompt the LLM to generate initialization scripts that set up the document with necessary elements and adjust their \texttt{selection} attributes according to the task description. 
% \ph{Talk about software specific prompts, following previous work}
% \ph{Could move this to Appendix if we need more space}
Following previous work with other virtual environments \cite{wang2024voyager}, we prompt the LLM to make the task code generic and reusable, as well as other Illustrator-specific instructions. The full prompt is shown in Appendix~\ref{app:prompt}. Table~\ref{tab:example} presents an example skill, including the corresponding task description, code implementation, and the layouts after running the initialization script and the skill, respectively.

We now detail how we operationalize our framework in Adobe Illustrator. We first execute the top-down simulation using the functionality categories listed in Appendix~\ref{app:func}, where the LLM explores tasks in each subcategory over three rounds. The validated skills, combined with sample scripts provided by the software, are then used to construct the API synergy graph. To train the GCN model for link prediction, we randomly split the sampled edges (both positive and negative) into training, development, and test sets with a ratio of 0.85/0.05/0.1.
The GCN model has two layers with an embedding size of $128$. We trained the model for $300$ epochs using the Adam optimizer with a learning rate of $0.01$.
Next, for each of the approximately 500 method APIs, we retrieve its top-$k$ ($k=5$) synergistic APIs and prompt the LLM to generate related tasks for one round.\footnote{We selected $k=5$ to balance between capturing potentially synergistic APIs and avoiding overwhelming the LLM with irrelevant API information.} 
In each round of the task creation stage (both top-down and bottom-up approaches), the LLM generates ten tasks. For each task, the LLM is allowed up to three trials with previous feedback to generate the code of skills.
We note that these hyperparameters may be subject to change based on cost considerations. We chose these values to ensure that the cost and size of the generated skillset remain manageable. 

\paragraph{Evaluation Setup}
% For the test set, 
To curate a test set that represents the needs of real users, we iteratively prompt the LLM to generate useful tasks in Illustrator. 
We manually verify the tasks and correct any issues in the corresponding initialization scripts, if any, to enable automatic evaluation, resulting in $94$ test tasks.
We use the \texttt{all-mpnet-base-v2} model from sentence BERT~\cite{reimers-2019-sentence-bert} to encode the task descriptions for both test set and the skillset built during offline simulation.
We use the test task description as the query and retrieve the relevant skills from the skillset with semantic matching through cosine similarity. 
% \ph{if we come to a more complex framework on inference time, we can add this to the method section.}

\paragraph{Baseline} The baseline method for this task is runtime generation where an LLM generates code to solve the query during runtime. We use \texttt{gpt-4o 2024-08-06}
as the LLM for baseline, as well as the ones in the proposed framework for fair comparisons. 
We note that advances in program synthesis methods are complementary to our approach; a stronger program synthesis model would benefit both the baseline and our framework.
To demonstrate the generalization ability of our framework, we also evaluated \texttt{lama-3.1-70B} and \texttt{deepseek-r1} as the code generation model on a random sample of $200$ tasks from each of the top-down and bottom-up strategies.

\paragraph{Proposed Methods}
(1) \textbf{Retrieval-Only (RO)}: Given a user query $q$, the system retrieves the most semantically relevant skill using cosine similarity over sentence embeddings. This approach ensures low-latency responses and minimal runtime cost, and avoids unsafe outputs, as all retrieved skills are pre-validated through offline trials. 
(2) \textbf{Retrieval-Augmented Generation (RAG)}: To explore the use of the skillset for handling arbitrary user queries, we prompt an LLM with the top-$r$ ($r=3$) retrieved skills as in-context examples. This enables flexible adaptation and composition beyond atomic tasks.

\paragraph{Evaluation Metrics}
We evaluate the generated skillset by assessing its ability to solve the given tasks. 
Specifically, we report the {\bf Success Rate}, which represents the proportion of tasks where an LVLM determines that the outcomes satisfy the provided task descriptions. 
This follows a similar procedure outlined in Section \ref{sec:code_gen}.
We further evaluate how reliable the LVLM's judgment is by comparing it with human judgment, the results are discussed in Section \ref{sec:lvlm_judge}.
We note that judging the actual outcome after execution is challenging. Previous work on LLM tool usage adopts a simpler setting, e.g., only focus on tasks in a question-answering format where answers can be easily verified~\cite{mialon2024gaia,yuan2024craft} or only checking the correctness of tool calling (without checking the execution results)~\cite{wang-etal-2024-llms-imaginarium}.
Additionally, because automating scripting interfaces requires consideration for user experiences and runtime cost, we report {\bf Response Time} as the averaged seconds per task for the system to output a script (either through retrieval or generation), along with averaged {\bf Token Cost} during runtime.
\section{Results and Discussions}
\label{sec:results}

\subsection{Effectiveness in Real-World Scenarios}
\label{sec:main}

Our proposed method demonstrates a substantial improvement over the baseline in automating tasks within Adobe Illustrator's scripting environment. 
As shown in Table~\ref{tab:main}, our approach (\textbf{RO}) achieves a success rate of $44.7\%$, outperforming the baseline's success rate of $28.7\%$ on the held-out test set. 
This evaluation setup effectively measures the end-to-end performance of automation systems when deployed in real-world software contexts, with the natural language task descriptions in the test set simulating real user queries.
In addition to RO, the RAG variant achieves a success rate of $42.6\%$, improving baseline performance by approximately $15\%$.
This demonstrates that retrieved skills not only serve as executable scripts but also act as effective in-context examples for code generation.

A key advantage of RO lies in its markedly lower runtime costs, which is a critical consideration given the potential volume and repetitive nature of user queries. 
RO avoids runtime code generation entirely by retrieving pre-validated skills, resulting in an average response time of just $0.1$ seconds for retrieval and zero token cost at runtime, as offline simulation is a one-time expense.
In contrast, both the baseline and RAG require significantly more resources at runtime, with response times of $4.0$ and $4.3$ seconds, and token costs of $666$ and $1219$.

% These results highlight the benefits of our method in terms of efficiency and resource utilization. By pre-generating a repository of verified skills, we enable rapid retrieval and execution of automation tasks without the latency and cost associated with real-time LLM interactions

% 27 Baseline

\begin{table}
\centering
\resizebox{\columnwidth}{!}{%
\begin{tabular}{lrrr} \toprule
         & Success Rate & Response Time & Token Cost  \\ \midrule
Baseline & 28.7\%       & 4.0 s          & 666         \\ \hdashline
{\bf RAG}      & 42.6\%       & 4.3 s          & 1219       \\
{\bf RO}     & 44.7\%       & 0.1 s          & 0*          \\ \bottomrule
\end{tabular}
}
\caption{Evaluation of the baseline and the proposed approaches. * The token cost for offline simulation in our approach is one-off and not included in the runtime cost.}
\label{tab:main}
\end{table}

\subsection{Effectiveness of Two Simulation Strategies}
\label{sec:sim_strategies}

\paragraph{Contribution to solving the test set.}
During offline simulation, we employed both top-down and bottom-up simulation strategies to capture a wide range of functionalities within the software's scripting environment.
Table \ref{tab:contribution} presents the contributions of each strategy and their combined impact when tackling tasks in the test set. 
Top-down simulation accounts for approximately $75\%$ of the skills retrieved.  
While the top-down strategy has a higher overall contribution, the bottom-up strategy still plays a significant role. 
It's important to note that the test set curation and the top-down simulation both involve prompting the LLM to generate useful tasks at a high level of abstraction based on the software's functionalities. This similarity potentially gives the top-down approach an advantage in matching test tasks. On the other hand, the bottom-up simulation, which prompts the LLM with low-level API information, is designed to cover long-tailed, less obvious skills.
Despite this, the bottom-up simulation still achieves comparable success rates relative to their contributions (top-down: $\frac{35.1\%}{75.5\%} = 46.5\%$, bottom-up: $\frac{9.8\%}{24.4\%} = 40.2\%$) to the test set that predominantly contains head-tail tasks — the most helpful tasks as determined by the LLM.

\begin{table}
\centering
\resizebox{\linewidth}{!}{%
\begin{tabular}{lrrr} \toprule
                   & Top-down & Bottom-up & Total   \\ \midrule
Successful Skills   & 35.1\%                   & 9.8\%          & 44.7\%  \\
Unsuccessful Skills & 40.4\%                   & 14.9\%          & 55.9\%  \\
Total              & 75.5\%                   & 24.4\%          & 100\%   \\ \bottomrule
\end{tabular}
}
\caption{Distribution of retrieved skills across op-down and bottom-up simulation strategies in the test set.}
\label{tab:contribution}
\end{table}

\paragraph{Bottom-up simulation uses more APIs but coverage remains incomplete.}
To further demonstrate that the bottom-up approach is effective in exploring the long-tail distribution of skills within the software's capabilities, Table \ref{tab:api_coverage} shows skills from bottom-up simulation cover 151 unique APIs, significantly more than the 49 APIs covered by the top-down approach and the 48 from sample scripts provided by the software.
This broader API coverage indicates that the bottom-up approach is successful in exploring a wide range of functionalities beyond the more commonly used APIs.

However, as mentioned in Section \ref{sec:task_create}, the bottom-up simulation iterated through approximately $378$ method APIs. The fact that only $151$ of these APIs resulted in successfully validated scripts suggests that many APIs did not have any corresponding successful tasks or scripts. We posit that the main reason for this limitation is the employed general-purpose LLMs' limited knowledge of Adobe Illustrator's extensive API library, resulting in LLMs' failure to generate relevant tasks or to produce code with correct API usage.

This outcome is expected, given the complexity and long-tail nature of software scripting API design. 
For example, a suite of APIs in Illustrator relates to physically printing, exporting to various file formats, and so on. 
These are ``end-of-workflow'' commands or meta-actions about the environment, rather than modular skills for manipulating document content.
Verifying their success is difficult in a closed-loop simulation. 
Some APIs are tied to specific legacy software versions.
Therefore, complete API coverage is an impractical goal.
However, the long tail still represents numerous specialized APIs that are infrequently used and therefore less likely to be well-represented in the LLMs' training data. This highlights room for future improvement, which we discuss in Section \ref{sec:limiations}.
% \ph{Add qualitative examples}

\begin{table}
\centering
\small
% \resizebox{\linewidth}{!}{%
\begin{tabular}{lrrr} \toprule
        & Sample Scripts & Top-down & Bottom-up  \\ \midrule
\# APIs & 48             & 49       & 151        \\ \bottomrule
\end{tabular}
% }
\caption{Number of API endpoints from Illustrator's inherit sample scripts and skills built from two simulation strategies.}
% \caption{Number of API endpoints in different sets of scripts/skills.}
\label{tab:api_coverage}
\end{table}

\begin{table}[]
    \centering
    \small
    \begin{tabular}{lrrr} \toprule
         &  Test Set  &  Top-down  &  Bottom-up \\ \midrule
    Avg. Score    &   2.48  &   2.28  & 1.75 \\ \bottomrule
    \end{tabular}
    \caption{Average usefulness scores (1–3 scale) of sampled tasks, as rated by human evaluators.}
    \label{tab:usefulness}
\end{table}

\paragraph{Human-perceived usefulness}
To assess the practical value of the discovered skills, we conducted a small-scale user study in which two Illustrator users with programming experience rated the usefulness of 150 sampled tasks on a 3-point scale: 1 (very rarely useful), 2 (occasionally helpful), and 3 (very useful). 
The sample included 50 tasks each from the test set, top-down skills, and bottom-up skills.
Annotators judged each task based solely on its description. 
% We adopted a stricter interpretation of the scale, acknowledging that most tasks offer some utility in niche scenarios, making it difficult to classify any task as completely useless.
We discuss the annotation setup and annotator agreement in Appendix \ref{app:help}.
As shown in Table~\ref{tab:usefulness}, skills produced by the top-down simulation strategy received a relatively high average usefulness score of $2.28$, closely matching the $2.48$ average for tasks in the test set. In contrast, bottom-up skills averaged $1.75$. Although this is relatively lower, it still leans toward the “occasionally helpful” category. This result is consistent with the bottom-up strategy’s goal of uncovering long-tail functionalities. We examine these differences further in the qualitative analysis below.

\paragraph{Qualitative Analysis} 
The top-down simulation tends to generate tasks that reflect high-level design intentions familiar to users. For instance, it produces skills such as ``arranging selected objects in a circle'' (Table~\ref{tab:example}), along with other layout patterns like zigzag or starburst formations. These tasks are aligned with common design workflows and are typically easy to interpret and apply.
The bottom-up simulation yields more technically specialized tasks that leverage less frequent APIs or scripting features. One example (Table~\ref{tab:full_example_2}) involves alerting the user when selected objects have tags that match a given list. Such skill introduces conditional logic and metadata inspection, which is expected to be rarer in practical usage but provides greater flexibility for advanced users.

\paragraph{Synergy Modeling of APIs in Bottom-up Simulation}
During bottom-up simulation, we use a GNN-based link prediction model to model the synergy of APIs. We show that this model can find synergistic APIs more effectively, compared to semantically matching similar APIs.
We evaluate the performance of the GCN model in retrieving synergistic APIs for task creation using the Hit@$k$ metric.
In this context, Hit@$k$ evaluates, for each API, how often the correct items are within the top-$k$ predicted items where correct items refer to APIs in existing skills (not in the training set).
We compare the link prediction model with a semantic matching baseline, which retrieves top-$k$ similar APIs using \texttt{all-mpnet-base-v2} embeddings and cosine similarity
Table \ref{tab:link_prediction} shows our model significantly outperforms semantic matching in identifying synergistic APIs, enhancing task creation in bottom-up simulation.

% \ph{TODO: we need to add an end-to-end evaluation with the semantic matching method, but is it worth it?}

\begin{table}
\centering
\footnotesize
\begin{tabular}{lrr} \toprule
      & Semantic matching & Link Prediction  \\ \midrule
Hit@5 & 16.7\%                         & 37.3\%                \\ \bottomrule
\end{tabular}
\caption{Comparison between the link prediction model and semantic matching method for identifying synergistic APIs. }
\label{tab:link_prediction}
\end{table}

\paragraph{Generation with trials is a key driver of the improved success rate for both simulation strategies.}
One key advantage of our offline simulation is the ability to perform multiple trials during code generation without incurring runtime penalties. We analyzed the impact of allowing up to three trials for code generation on the overall success rate.
Table \ref{tab:trials} demonstrates that the success rate nearly doubles after three trials for both top-down and bottom-up strategies. Specifically, the success rate for both top-down and bottom-up strategies nearly doubles after three trials.
We found similar improvement when using \texttt{Llama-3.1-70B} and \texttt{deepseek-r1} as the code generation models shown in Table~\ref{tab:model_strategy_success}.

\begin{table}
\centering
\resizebox{\linewidth}{!}{%
\begin{tabular}{lrrr} \toprule
          & \# Tasks & \%Success@1 & \%Success@3  \\ \midrule
Top-down  & 1721     & 16.7\%      & 34.9\%                           \\
Bottom-up & 3256     & 23.1\%      & 46.6\%                           \\ \bottomrule
\end{tabular}
}
\caption{Success rate at the first and third trials during simulations.}
\label{tab:trials}
\end{table}

\subsection{Reliability of LVLM's judgment}
\label{sec:lvlm_judge}
% Assessing the success of generated scripts requires reliable verification.
We employed an LVLM to evaluate whether the execution outcomes satisfy the task descriptions. To estimate the reliability of the LVLM's judgments, we compared its assessments with human evaluations on a sample of $122$ tasks. 
Table \ref{tab:LVLM} shows that the LVLM has a precision of $90.9\%$ and a recall of $80.0\%$. Out of $50$ tasks deemed successful by humans, the LVLM correctly identified $40$ as successful and $10$ as failures. Meanwhile, the LVLM only incorrectly labeled $4$ tasks as successful when they were not (false positives). These results indicate that the LVLM is generally reliable, albeit slightly conservative in its judgments.

\begin{table}[t]
\centering
\resizebox{\linewidth}{!}{%
\begin{tabular}{lrr} \toprule
                                 & Success (LVLM) & Fail (LVLM) \\ \midrule
Success (Human)                  & 40 (TP)                           & 10 (FN)                          \\
Fail (Human) & 4 (FP)                            & 68 (TN)                         \\ \bottomrule
\end{tabular}
}
\caption{Confusion matrix between human and LVLM judgments.}
\label{tab:LVLM}
\end{table}

\section{Conclusions}

We proposed an offline simulation framework for skill discovery in software scripting automation with LLMs. Our method pre-build a diverse and verified skillset in two phases: (1) task creation, guided by both top-down functionality categories and bottom-up API synergy exploration, and (2) skill generation with trials, where LLM-generated scripts are iteratively refined through feedback based on execution error and functional outcome.

Experiments on Adobe Illustrator demonstrated that our method achieves higher success rates, broader functional coverage, and lower response time compared to runtime generation. The top-down task creation strategy encourages broad functional coverage, while the bottom-up API synergy modeling explores the of discovered skills, particularly for underutilized API endpoints. Furthermore, our GCN-based API link prediction model outperformed naive semantic matching in identifying synergistic API pairs, enhancing task creation during the bottom-up simulation.

To our best knowledge, this is the first attempt to use software scripting interfaces as a testbed for LLM-based systems, highlighting the advantage of obtaining direct execution feedback in a controlled environment and providing valuable insights into the design of more capable and user-aligned scripting systems.

\section*{Limitations}
\label{sec:limiations}
Despite the promising results of our offline simulation framework for skill discovery in software scripting automation, there are several limitations.

First, our method relies heavily on publicly available scripting guides and the existing knowledge of LLMs about the software. In cases where such documentation is incomplete, outdated, or unavailable, or when LLMs have limited knowledge of the software, LLMs may hallucinate API usage or rely on superficial cues~\cite{patil2024gorilla,zhou2024explore,liu2025large}, reducing effectiveness and robustness. 
This dependence restricts the applicability of our framework to well-documented software systems and those familiar to the LLMs. 
To address this, one promising direction is to train a domain-specific LLM tailored to the software through iterative trial and error, which could generate more accurate and comprehensive skills.
Such methods are complementary to our framework.

Second, our evaluation primarily focuses on automated success rates and performance metrics within a controlled environment. While these metrics provide valuable insights into technical performance, they do not capture user experience or satisfaction. Automatic evaluation can be less reliable on such subjective or high-inference criteria~\cite{liu-etal-2016-evaluate,amershi2019guidelines,xu-etal-2024-promises}. 
Although we ran a small-scale user study to evaluate the perceived usefulness of the explored tasks,
future studies could include more comprehensive user evaluations to assess the usability and overall impact of the generated skills in real-world workflows. 
Additionally, mining API interactions from actual scripting projects could provide valuable insights into common user behaviors and task patterns, making the generated skills more applicable in practice.
However, obtaining such data is challenging and raises interesting questions about what kinds of behavioral data are useful and how to collect them responsibly.

Third, we expect that queries from real users may be more complex and not always match the granularity of the skills in our skillset. Real-world user queries might be ambiguous or involve higher-level tasks that require combining multiple skills or adapting existing ones. The skillset generated by our framework does not account for individual user preferences or specific workflow requirements. 
However, the proposed RAG variant shows promising results, significantly improving the success rate with only a modest increase in response time, suggesting its potential to better handle such complex or personalized queries.

Fourth, due to limited support for external programmatic control, we are only able to conduct a large-scale study on Adobe Illustrator. Although we conducted initial exploration on Excel, future work should extend the framework to other scripting environments with broader automation support to validate its generalizability.

\section*{Acknowledgments}
We would like to thank our collaborators at Adobe Research and the University of Maryland for their helpful discussion, including Dang Minh Nguyen, Dayeon Ki, Nishant Deepa Balepur, Vishakh Padmakumar, Yoonjoo Lee, and Hyunji Lee.
\bibliography{custom}

@article{liu2025large,
  title={Large language models and causal inference in collaboration: A comprehensive survey},
  author={Liu, Xiaoyu and Xu, Paiheng and Wu, Junda and Yuan, Jiaxin and Yang, Yifan and Zhou, Yuhang and Liu, Fuxiao and Guan, Tianrui and Wang, Haoliang and Yu, Tong and others},
  journal={Findings of the Association for Computational Linguistics: NAACL 2025},
  pages={7668--7684},
  year={2025}
}

@inproceedings{xu-etal-2024-promises,
    title = "The Promises and Pitfalls of Using Language Models to Measure Instruction Quality in Education",
    author = "Xu, Paiheng  and
      Liu, Jing  and
      Jones, Nathan  and
      Cohen, Julie  and
      Ai, Wei",
    editor = "Duh, Kevin  and
      Gomez, Helena  and
      Bethard, Steven",
    booktitle = "Proceedings of the 2024 Conference of the North American Chapter of the Association for Computational Linguistics: Human Language Technologies (Volume 1: Long Papers)",
    month = jun,
    year = "2024",
    address = "Mexico City, Mexico",
    publisher = "Association for Computational Linguistics",
    url = "https://aclanthology.org/2024.naacl-long.246/",
    doi = "10.18653/v1/2024.naacl-long.246",
    pages = "4375--4389"
}

@inproceedings{amershi2019guidelines,
  title={Guidelines for human-AI interaction},
  author={Amershi, Saleema and Weld, Dan and Vorvoreanu, Mihaela and Fourney, Adam and Nushi, Besmira and Collisson, Penny and Suh, Jina and Iqbal, Shamsi and Bennett, Paul N and Inkpen, Kori and others},
  booktitle={Proceedings of the 2019 chi conference on human factors in computing systems},
  pages={1--13},
  year={2019}
}

@inproceedings{liu-etal-2016-evaluate,
    title = "How {NOT} To Evaluate Your Dialogue System: An Empirical Study of Unsupervised Evaluation Metrics for Dialogue Response Generation",
    author = "Liu, Chia-Wei  and
      Lowe, Ryan  and
      Serban, Iulian  and
      Noseworthy, Mike  and
      Charlin, Laurent  and
      Pineau, Joelle",
    editor = "Su, Jian  and
      Duh, Kevin  and
      Carreras, Xavier",
    booktitle = "Proceedings of the 2016 Conference on Empirical Methods in Natural Language Processing",
    month = nov,
    year = "2016",
    address = "Austin, Texas",
    publisher = "Association for Computational Linguistics",
    url = "https://aclanthology.org/D16-1230/",
    doi = "10.18653/v1/D16-1230",
    pages = "2122--2132"
}

@inproceedings{zhou2024explore,
  title={Explore spurious correlations at the concept level in language models for text classification},
  author={Zhou, Yuhang and Xu, Paiheng and Liu, Xiaoyu and An, Bang and Ai, Wei and Huang, Furong},
  booktitle={Proceedings of the 62nd Annual Meeting of the Association for Computational Linguistics (Volume 1: Long Papers)},
  pages={478--492},
  year={2024}
}

@article{patil2024gorilla,
  title={Gorilla: Large language model connected with massive apis},
  author={Patil, Shishir G and Zhang, Tianjun and Wang, Xin and Gonzalez, Joseph E},
  journal={Advances in Neural Information Processing Systems},
  volume={37},
  pages={126544--126565},
  year={2024}
}

@article{wu2022graph,
  title={Graph neural networks in recommender systems: a survey},
  author={Wu, Shiwen and Sun, Fei and Zhang, Wentao and Xie, Xu and Cui, Bin},
  journal={ACM Computing Surveys},
  volume={55},
  number={5},
  pages={1--37},
  year={2022},
  publisher={ACM New York, NY}
}

@article{xu2025gfairhint,
  title={Gfairhint: Improving individual fairness for graph neural networks via fairness hint},
  author={Xu, Paiheng and Zhou, Yuhang and An, Bang and Ai, Wei and Huang, Furong},
  journal={ACM Transactions on Knowledge Discovery from Data},
  volume={19},
  number={3},
  pages={1--22},
  year={2025},
  publisher={ACM New York, NY}
}

@inproceedings{nguyen2025gui,
  title={Gui agents: A survey},
  author={Nguyen, Dang and Chen, Jian and Wang, Yu and Wu, Gang and Park, Namyong and Hu, Zhengmian and Lyu, Hanjia and Wu, Junda and Aponte, Ryan and Xia, Yu and others},
  booktitle={Findings of the Association for Computational Linguistics: ACL 2025},
  pages={22522--22538},
  year={2025}
}

@article{liu2024explore,
  title={Explore then Determine: A GNN-LLM Synergy Framework for Reasoning over Knowledge Graph},
  author={Liu, Guangyi and Zhang, Yongqi and Li, Yong and Yao, Quanming},
  journal={arXiv preprint arXiv:2406.01145},
  year={2024}
}

@article{huang20221+,
  title={1+ 1> 2: Programming know-what and know-how knowledge fusion, semantic enrichment and coherent application},
  author={Huang, Qing and Yuan, Zhiqiang and Xing, Zhenchang and Zuo, Zhengkang and Wang, Changjing and Xia, Xin},
  journal={IEEE Transactions on Services Computing},
  volume={16},
  number={3},
  pages={1540--1554},
  year={2022},
  publisher={IEEE}
}

@article{qi2022correlation,
  title={A correlation graph based approach for personalized and compatible web apis recommendation in mobile app development},
  author={Qi, Lianyong and Lin, Wenmin and Zhang, Xuyun and Dou, Wanchun and Xu, Xiaolong and Chen, Jinjun},
  journal={IEEE Transactions on Knowledge and Data Engineering},
  volume={35},
  number={6},
  pages={5444--5457},
  year={2022},
  publisher={IEEE}
}

@article{anokhin2024arigraph,
  title={Arigraph: Learning knowledge graph world models with episodic memory for llm agents},
  author={Anokhin, Petr and Semenov, Nikita and Sorokin, Artyom and Evseev, Dmitry and Burtsev, Mikhail and Burnaev, Evgeny},
  journal={arXiv preprint arXiv:2407.04363},
  year={2024}
}

@article{deng2024mind2web,
  title={Mind2web: Towards a generalist agent for the web},
  author={Deng, Xiang and Gu, Yu and Zheng, Boyuan and Chen, Shijie and Stevens, Sam and Wang, Boshi and Sun, Huan and Su, Yu},
  journal={Advances in Neural Information Processing Systems},
  volume={36},
  year={2024}
}

@article{ma2024agentboard,
  title={AgentBoard: An Analytical Evaluation Board of Multi-turn LLM Agents},
  author={Ma, Chang and Zhang, Junlei and Zhu, Zhihao and Yang, Cheng and Yang, Yujiu and Jin, Yaohui and Lan, Zhenzhong and Kong, Lingpeng and He, Junxian},
  journal={arXiv preprint arXiv:2401.13178},
  year={2024}
}

@article{yao2022webshop,
  title={Webshop: Towards scalable real-world web interaction with grounded language agents},
  author={Yao, Shunyu and Chen, Howard and Yang, John and Narasimhan, Karthik},
  journal={Advances in Neural Information Processing Systems},
  volume={35},
  pages={20744--20757},
  year={2022}
}

@inproceedings{gao2024assistgui,
  title={AssistGUI: Task-Oriented PC Graphical User Interface Automation},
  author={Gao, Difei and Ji, Lei and Bai, Zechen and Ouyang, Mingyu and Li, Peiran and Mao, Dongxing and Wu, Qinchen and Zhang, Weichen and Wang, Peiyi and Guo, Xiangwu and others},
  booktitle={Proceedings of the IEEE/CVF Conference on Computer Vision and Pattern Recognition},
  pages={13289--13298},
  year={2024}
}

@article{li2024sheetcopilot,
  title={SheetCopilot: Bringing software productivity to the next level through large language models},
  author={Li, Hongxin and Su, Jingran and Chen, Yuntao and Li, Qing and ZHANG, ZHAO-XIANG},
  journal={Advances in Neural Information Processing Systems},
  volume={36},
  year={2024}
}

@inproceedings{ma2024spreadsheetbench,
  title={SpreadsheetBench: Towards Challenging Real World Spreadsheet Manipulation},
  author={Ma, Zeyao and Zhang, Bohan and Zhang, Jing and Yu, Jifan and Zhang, Xiaokang and Zhang, Xiaohan and Luo, Sijia and Wang, Xi and Tang, Jie},
  booktitle={The Thirty-eight Conference on Neural Information Processing Systems Datasets and Benchmarks Track},
  year={2024}
}

@article{cao2024spider2,
  title={Spider2-V: How Far Are Multimodal Agents From Automating Data Science and Engineering Workflows?},
  author={Cao, Ruisheng and Lei, Fangyu and Wu, Haoyuan and Chen, Jixuan and Fu, Yeqiao and Gao, Hongcheng and Xiong, Xinzhuang and Zhang, Hanchong and Mao, Yuchen and Hu, Wenjing and others},
  journal={arXiv preprint arXiv:2407.10956},
  year={2024}
}

@article{xie2024osworld,
  title={Osworld: Benchmarking multimodal agents for open-ended tasks in real computer environments},
  author={Xie, Tianbao and Zhang, Danyang and Chen, Jixuan and Li, Xiaochuan and Zhao, Siheng and Cao, Ruisheng and Hua, Toh Jing and Cheng, Zhoujun and Shin, Dongchan and Lei, Fangyu and others},
  journal={arXiv preprint arXiv:2404.07972},
  year={2024}
}

@inproceedings{qian2023creator,
  title={CREATOR: Tool Creation for Disentangling Abstract and Concrete Reasoning of Large Language Models},
  author={Qian, Cheng and Han, Chi and Fung, Yi and Qin, Yujia and Liu, Zhiyuan and Ji, Heng},
  booktitle={Findings of the Association for Computational Linguistics: EMNLP 2023},
  pages={6922--6939},
  year={2023}
}

@article{nguyen2024dynasaur,
  title={DynaSaur: Large Language Agents Beyond Predefined Actions},
  author={Nguyen, Dang and Lai, Viet Dac and Yoon, Seunghyun and Rossi, Ryan A and Zhao, Handong and Zhang, Ruiyi and Mathur, Puneet and Lipka, Nedim and Wang, Yu and Bui, Trung and others},
  journal={arXiv preprint arXiv:2411.01747},
  year={2024}
}

@inproceedings{cai2024large,
  title={Large Language Models as Tool Makers},
  author={Cai, Tianle and Wang, Xuezhi and Ma, Tengyu and Chen, Xinyun and Zhou, Denny},
  booktitle={The Twelfth International Conference on Learning Representations},
  year={2024}
}

@article{zhao2024agentic,
  title={Agentic Skill Discovery},
  author={Zhao, Xufeng and Weber, Cornelius and Wermter, Stefan},
  journal={arXiv preprint arXiv:2405.15019},
  year={2024}
}

@article{pan2024automatically,
  title={Automatically correcting large language models: Surveying the landscape of diverse automated correction strategies},
  author={Pan, Liangming and Saxon, Michael and Xu, Wenda and Nathani, Deepak and Wang, Xinyi and Wang, William Yang},
  journal={Transactions of the Association for Computational Linguistics},
  volume={12},
  pages={484--506},
  year={2024},
  publisher={MIT Press One Broadway, 12th Floor, Cambridge, Massachusetts 02142, USA~…}
}

@article{kim2024language,
  title={Language models can solve computer tasks},
  author={Kim, Geunwoo and Baldi, Pierre and McAleer, Stephen},
  journal={Advances in Neural Information Processing Systems},
  volume={36},
  year={2024}
}

@article{zhang2023unifying,
  title={Unifying the perspectives of nlp and software engineering: A survey on language models for code},
  author={Zhang, Ziyin and Chen, Chaoyu and Liu, Bingchang and Liao, Cong and Gong, Zi and Yu, Hang and Li, Jianguo and Wang, Rui},
  journal={arXiv preprint arXiv:2311.07989},
  year={2023}
}

@article{austin2021program,
  title={Program synthesis with large language models},
  author={Austin, Jacob and Odena, Augustus and Nye, Maxwell and Bosma, Maarten and Michalewski, Henryk and Dohan, David and Jiang, Ellen and Cai, Carrie and Terry, Michael and Le, Quoc and others},
  journal={arXiv preprint arXiv:2108.07732},
  year={2021}
}

@inproceedings{nijkamp2023codegen,
  title={CodeGen: An Open Large Language Model for Code with Multi-Turn Program Synthesis},
  author={Nijkamp, Erik and Pang, Bo and Hayashi, Hiroaki and Tu, Lifu and Wang, Huan and Zhou, Yingbo and Savarese, Silvio and Xiong, Caiming},
  booktitle={The Eleventh International Conference on Learning Representations},
  year={2023}
}

@article{liu2024your,
  title={Is your code generated by chatgpt really correct? rigorous evaluation of large language models for code generation},
  author={Liu, Jiawei and Xia, Chunqiu Steven and Wang, Yuyao and Zhang, Lingming},
  journal={Advances in Neural Information Processing Systems},
  volume={36},
  year={2024}
}

@article{li2022competition,
  title={Competition-level code generation with alphacode},
  author={Li, Yujia and Choi, David and Chung, Junyoung and Kushman, Nate and Schrittwieser, Julian and Leblond, R{\'e}mi and Eccles, Tom and Keeling, James and Gimeno, Felix and Dal Lago, Agustin and others},
  journal={Science},
  volume={378},
  number={6624},
  pages={1092--1097},
  year={2022},
  publisher={American Association for the Advancement of Science}
}

@inproceedings{yuan2024craft,
  title={CRAFT: Customizing LLMs by Creating and Retrieving from Specialized Toolsets},
  author={Yuan, Lifan and Chen, Yangyi and Wang, Xingyao and Fung, Yi and Peng, Hao and Ji, Heng},
  booktitle={The Twelfth International Conference on Learning Representations},
  year={2024}
}

@inproceedings{mialon2024gaia,
  title={GAIA: a benchmark for General AI Assistants},
  author={Mialon, Gr{\'e}goire and Fourrier, Cl{\'e}mentine and Wolf, Thomas and LeCun, Yann and Scialom, Thomas},
  booktitle={The Twelfth International Conference on Learning Representations},
  year={2024}
}

@inproceedings{reimers-2019-sentence-bert,
  title = "Sentence-BERT: Sentence Embeddings using Siamese BERT-Networks",
  author = "Reimers, Nils and Gurevych, Iryna",
  booktitle = "Proceedings of the 2019 Conference on Empirical Methods in Natural Language Processing",
  month = "11",
  year = "2019",
  publisher = "Association for Computational Linguistics",
  url = "https://arxiv.org/abs/1908.10084",
}

@inproceedings{kipf2017semi,
  title={Semi-supervised classification with graph convolutional networks},
  author={Kipf, Thomas N and Welling, Max},
  booktitle={5th International Conference on Learning Representations (ICLR)},
  year={2017},
  url={https://arxiv.org/abs/1609.02907}
}

@inproceedings{wang-etal-2024-llms-imaginarium,
    title = "{LLM}s in the Imaginarium: Tool Learning through Simulated Trial and Error",
    author = "Wang, Boshi  and
      Fang, Hao  and
      Eisner, Jason  and
      Van Durme, Benjamin  and
      Su, Yu",
    editor = "Ku, Lun-Wei  and
      Martins, Andre  and
      Srikumar, Vivek",
    booktitle = "Proceedings of the 62nd Annual Meeting of the Association for Computational Linguistics (Volume 1: Long Papers)",
    month = aug,
    year = "2024",
    address = "Bangkok, Thailand",
    publisher = "Association for Computational Linguistics",
    url = "https://aclanthology.org/2024.acl-long.570",
    doi = "10.18653/v1/2024.acl-long.570",
    pages = "10583--10604",
}

@article{wang2024voyager,
  title={Voyager: An Open-Ended Embodied Agent with Large Language Models},
  author={Wang, Guanzhi and Xie, Yuqi and Jiang, Yunfan and Mandlekar, Ajay and Xiao, Chaowei and Zhu, Yuke and Fan, Linxi and Anandkumar, Anima},
  journal={Transactions on Machine Learning Research},
  year={2024}
}

@article{bubeck2023sparks,
  title={Sparks of artificial general intelligence: Early experiments with gpt-4},
  author={Bubeck, S{\'e}bastien and Chandrasekaran, Varun and Eldan, Ronen and Gehrke, Johannes and Horvitz, Eric and Kamar, Ece and Lee, Peter and Lee, Yin Tat and Li, Yuanzhi and Lundberg, Scott and others},
  journal={arXiv preprint arXiv:2303.12712},
  year={2023}
}

@inproceedings{zhao-etal-2024-nl2formula,
    title = "{NL}2{F}ormula: Generating Spreadsheet Formulas from Natural Language Queries",
    author = "Zhao, Wei  and
      Hou, Zhitao  and
      Wu, Siyuan  and
      Gao, Yan  and
      Dong, Haoyu  and
      Wan, Yao  and
      Zhang, Hongyu  and
      Sui, Yulei  and
      Zhang, Haidong",
    editor = "Graham, Yvette  and
      Purver, Matthew",
    booktitle = "Findings of the Association for Computational Linguistics: EACL 2024",
    month = mar,
    year = "2024",
    address = "St. Julian{'}s, Malta",
    publisher = "Association for Computational Linguistics",
    url = "https://aclanthology.org/2024.findings-eacl.158",
    pages = "2377--2388",
}

@article{chen2021evaluating,
  title={Evaluating large language models trained on code},
  author={Chen, Mark and Tworek, Jerry and Jun, Heewoo and Yuan, Qiming and Pinto, Henrique Ponde De Oliveira and Kaplan, Jared and Edwards, Harri and Burda, Yuri and Joseph, Nicholas and Brockman, Greg and others},
  journal={arXiv preprint arXiv:2107.03374},
  year={2021}
}

@article{ousterhout1998scripting,
  title={Scripting: Higher level programming for the 21st century},
  author={Ousterhout, John K},
  journal={Computer},
  volume={31},
  number={3},
  pages={23--30},
  year={1998},
  publisher={IEEE}
}

@article{gandhi2023natural,
  title={Natural Language Commanding via Program Synthesis},
  author={Gandhi, Apurva and Nguyen, Thong Q and Jiao, Huitian and Steen, Robert and Bhatawdekar, Ameya},
  journal={arXiv preprint arXiv:2306.03460},
  year={2023}
}

\appendix

\section{Adobe Illustrator Functionalities}
\label{app:func}

Part of the functionality categories in Illustrator are shown in Table \ref{tab:illustrator_features}.

\section{Prompts}
\label{app:prompt}

In this section, we show system prompts and user prompts for task creation, skill generation with trials, and LVLM-based validation in Table \ref{prompt:creation}, Table \ref{prompt:sys_code}, and Table \ref{prompt:validation}, respectively.

\section{Example Illustrator Skills}
\label{app:example}
We show an example skill generated by the top-down strategy in Table \ref{tab:full_example}, and one by the bottom-up strategy in Table \ref{tab:full_example_2}.

\section{Annotation for Perceived Helpfulness}
\label{app:help}

We asked the two annotators to rate the usefulness of 150 sampled tasks on a 3-point scale: 1 (very rarely useful), 2 (occasionally helpful), and 3 (very useful). Annotators judged each task based solely on its description, and the tasks are randomly shuffled without disclosing 
We adopted a stricter interpretation of the scale, acknowledging that most tasks offer some utility in niche scenarios, making it difficult to classify any task as completely useless.

The annotators achieved substantial agreement on their ratings, with similar average scores across conditions and only minor differences between them, as shown in Table~\ref{tab:usefulness_full}.
While exact agreement occurred in $41\%$ of cases, the one-off agreement was remarkably high at $95\%$, indicating that when disagreements occurred, they typically differed by only one point on the scale. 
This is further supported by the weighted Cohen's Kappa score of $0.176$, which accounts for the ordinal nature of our rating scale and the fact that most disagreements were minor. 
Krippendorff's Alpha (0.171) similarly reflects this pattern of agreement. 
Overall, these metrics demonstrate that despite the inherent subjectivity in usefulness judgments, our annotators maintained reasonable consistency in their evaluations.

\section{Exploring Excel as the Testbed}
\label{app:excel}

\begin{table}[]
    \centering
    \small
    \begin{tabular}{lrrr} \toprule
         &  Test Set  &  Top-down  &  Bottom-up \\ \midrule
    Rater 1    &   2.42  &   2.40  & 1.78 \\ 
    Rater 2    &   2.53  &   2.16  & 1.72 \\ \bottomrule
    \end{tabular}
    \caption{Average usefulness scores (1–3 scale) of sampled tasks, as rated by human evaluators.}
    \label{tab:usefulness_full}
\end{table}

\begin{table}[t]
    \centering
    \footnotesize
    \begin{tabular}{lr} \toprule
    Task     &  \# Trials \\ \midrule
    Combine multiple Excel tables into one     &  5 \\
    Count blank rows on all sheets   & Failed \\
    Return table data as JSON  & Failed \\
    Remove hyperlinks from each cell & 5 \\
    Move rows using range values  & 1* \\ \bottomrule
    \end{tabular}
    \caption{Number of LLM code generation trials required to successfully complete each Excel scripting task. “Failed” indicates that the LLM was unable to produce a working script for the task within five attempts. * due to the ambiguity in task description, it didn't reproduce the example code in the tutorial but was able to achieve something reasonable.}
    \label{tab:excel}
\end{table}

\begin{table}[t]
\centering
\resizebox{\linewidth}{!}{%
\begin{tabular}{llrr} \toprule
Model & Strategy & \%Success@1 & \multicolumn{1}{c}{\%Success@3} \\ \midrule
Llama-3.1-70B & Top-down  & 5.5\% & 16.8\% \\
Llama-3.1-70B & Bottom-up & 8.7\% & 23.5\% \\
deepseek-r1   & Top-down  & 6.9\% & 27.6\% \\
deepseek-r1   & Bottom-up & 8.7\% & 34.8\% \\ \bottomrule
\end{tabular}
}
\caption{Success rate at the first and third trials for each model and strategy.}
\label{tab:model_strategy_success}
\end{table}

While Adobe Illustrator provides a flexible scripting interface that allows external automation of tasks, Excel’s scripting support is primarily designed for use within its own environment and does not readily permit programmatic control from outside applications. As a result, the approach used to automate and evaluate scripts in Illustrator could not be directly applied to Excel. This limitation restricted our ability to conduct comparable experiments across both platforms.

However, we explore the following aspects to generalize our framework to Excel.
(1) {\bf API analysis}. We collected 2,140 APIs and identified 286 sample scripts from the official tutorial, \footnote{https://learn.microsoft.com/en-us/office/dev/scripts/overview/excel}
which cover 124 unique APIs. This supports our broader observation that modern software exposes extensive scripting APIs, while available example scripts typically cover only a small subset.
(2) {\bf LLM-based code generation}. We selected five tasks from the ``quick scenarios'' category in the Excel scripting tutorial and prompted LLM to generate scripts following our framework (see Section~\ref{sec:code_gen}). We show the system prompt for Excel in Table \ref{prompt:sys_code_excel} and user prompt follows the one in Table \ref{prompt:sys_code}. These tasks are aligned with the complexity and skillset expected in our Illustrator experiments.
Table~\ref{tab:excel} reports the number of LLM code generation trials required for each task.
We observed that some tasks could be completed with only a few LLM prompt iterations, while others consistently failed, likely due to differences in task difficulty and model familiarity with Excel’s scripting environment. Importantly, two out of five tasks require multiple trials highlights the effectiveness of our offline simulation framework and the trial-and-error approach. Overall, these results indicate both the challenges and opportunities in generalizing LLM-based code generation across software platforms with distinct automation interfaces, and demonstrate that our framework can be extended beyond its initial domain.

\begin{table}[!htb]
    \centering
    \footnotesize
    \renewcommand{\arraystretch}{1.5}
    \begin{tabular}{p{0.13\linewidth} p{0.8\linewidth}}
    % \begin{tabular}{ >{\centering\arraybackslash}m{0.15\columnwidth} p{0.85\columnwidth}}
    % \begin{tabular}{cc}
        \toprule
        Desc & arrange selected objects in circle  \\ \hline
        \# Trial & No. 2 \\ \hline
        Code last round (w/ Init. code) &  \begin{minipage}[t]{1\linewidth}
        \vspace{-\baselineskip}
        \begin{lstlisting}[language=java, basicstyle=\ttfamily\tiny, backgroundcolor=\color{gray!10}, keywordstyle=\color{blue}, commentstyle=\color{green}, stringstyle=\color{red}]
function createSimpleLayout() {
    var doc = app.documents.add();
    var numObjs = 5;
    var radius = 100;
    for (var i = 0; i < numObjs; i++) {
        var circle = doc.pathItems.ellipse(...);
        circle.filled = true;
        circle.fillColor = doc.colors.add();
        circle.fillColor.red = i * 50;
        circle.selected = true;
    }
    return 'Initialization complete';
}
createSimpleLayout();

function arrangeInCircle(cX, cY, radius) {
    var sel = app.selection;
    if (sel.length === 0) {
        throw new Error('No objects selected');
    }
    var angleStep = 360 / sel.length;
    for (var i = 0; i < sel.length; i++) {
        var angle = angleStep*i*(Math.PI / 180);
        var x = cX + radius * Math.cos(angle);
        var y = cY + radius * Math.sin(angle);
        sel[i].position = [x, y];
    }
    return 'Objects arranged in a circle';
}

arrangeInCircle(300, 300, 150);\end{lstlisting}
        \end{minipage} \\ \hline
        Error msg  &  Adobe Illustrator got an error: Error 21: undefined is not an object.
Line: 17
->          circle.fillColor = doc.colors.add(); \\ \hline
        Judge & {\it Not applicable because code failed to run} \\ \hline
        Init. code & \begin{minipage}[t]{1\linewidth}
        \vspace{-\baselineskip}
        \begin{lstlisting}[language=java, basicstyle=\ttfamily\tiny, backgroundcolor=\color{gray!10}, keywordstyle=\color{blue}, commentstyle=\color{green}, stringstyle=\color{red}]
function createSimpleLayout() {
    var doc = app.documents.add();
    var numObjs = 5;
    var radius = 100;
    for (var i = 0; i < numObjs; i++) {
        var circle = doc.pathItems.ellipse(...);
        circle.filled = true;
        var color = new RGBColor();
        color.red = i * 50;
        circle.fillColor = color;
        circle.selected = true;
    }
    return 'Initialization complete';
}

createSimpleLayout();\end{lstlisting}
        \end{minipage}
        \\ \hline
        Code & \begin{minipage}[t]{0.9\linewidth}
        \vspace{-\baselineskip}
        \begin{lstlisting}[language=java, basicstyle=\ttfamily\tiny, backgroundcolor=\color{gray!10}, keywordstyle=\color{blue}, commentstyle=\color{green}, stringstyle=\color{red}]
function arrangeInCircle(cX, cY, radius) {
  var sel = app.selection
  if (sel.length === 0) {
    throw new Error('No selection');
    }
  var angleStep = 360 / sel.length;
  for (var i = 0; i < sel.length; i++) {
    var angle = angleStep*i*(Math.PI/180);
    var x = cX + radius * Math.cos(angle);
    var y = cY + radius * Math.sin(angle);
    sel[i].position = [x, y];
    }
  return 'Objects arranged in a circle';
}
        \end{lstlisting}
        \end{minipage}
        \\ \hline
        Effect & \includegraphics[width=0.1\linewidth]{figures/init.png} $\Rightarrow$ \includegraphics[width=0.2\linewidth]{figures/outcome.png} \\
        \bottomrule
    \end{tabular}
    \caption{Complete version of Table \ref{tab:example}. \texttt{Desc} is the natural language description. \texttt{Effect} shows the layout before and after running the skill. \texttt{Judge} is the validation judgment of the LVLM. Code simplified for space constraints.}
    \label{tab:full_example}
\end{table}

\begin{table}[!htb]
    \centering
    \footnotesize
    \renewcommand{\arraystretch}{1.5}
    \begin{tabular}{p{0.13\linewidth} p{0.8\linewidth}}
    % \begin{tabular}{ >{\centering\arraybackslash}m{0.15\columnwidth} p{0.85\columnwidth}}
    % \begin{tabular}{cc}
        \toprule
        Desc & Alert the user when the name of a tag in the selection matches a predefined list of names.  \\ \hline
        \# Trial & No. 2 \\ \hline
        Code last round (w/ Init. code) &  \begin{minipage}[t]{1.02\linewidth}
        \vspace{-\baselineskip}
        \begin{lstlisting}[language=java, basicstyle=\ttfamily\tiny, backgroundcolor=\color{gray!10}, keywordstyle=\color{blue}, commentstyle=\color{ForestGreen}, stringstyle=\color{red}]
var doc = app.documents.add();

// Init code
var textFrame = doc.textFrames.add();
textFrame.contents = "Sample Text";
textFrame.name = "Tag1";

var rect = doc.pathItems.rectangle(...);
rect.name = "Tag2";

var selectionItems = [textFrame, rect];
for (var i = 0; i < selectionItems.length; i++) {
	selectionItems[i].selected = true;
}

// Task code
function alertIfTagNameMatches(predefined) {
    try {
        var items = app.activeDocument.selection;
        if (items.length === 0) {
            throw new Error("No items selected.");
        }
        var matchedTags = [];
        for (var i = 0; i < items.length; i++) {
            var item = items[i].name;
            if (predefined.indexOf(item)!==-1) {
                matchedTags.push(item);
            }
        }
        if (matchedTags.length > 0) {
            return matchedTags.join(", ");
        } else {
            return "No matched tags found.";
        }
    } catch (e) {
        return "Error: " + e.message;
    }
}

var predefinedNames = ["Tag1", "Tag2", "Tag3"];
alertIfTagNameMatches(predefinedNames);\end{lstlisting}
        \end{minipage} \\ \hline
        Error msg  &  {\it Not applicable because the code was successfully executed.} \\ \hline
        Judge & The task requires alerting the user if the tag names match any in the predefined list, but the task code only returns the result as a string. There is no actual alert mechanism used. Replace the return statements with an alert function to notify the user, e.g., alert(``Matched tag(s): '' + matchedTags.join(``, '')); to fulfill the requirement correctly. \\ \hline
        Init. code & {\it Same as above}
        \\ \hline
        Code & {\it Replaced return matchedTags.join(``, ''); with alert(``Matched tag(s):'' + matchedTags.join(``, ''));}\\ \hline
        Effect & {\it An alert message box is popped up.} \\
        \bottomrule
    \end{tabular}
    \caption{Example script generated by bottom-up strategy. Details simplified for space constraints. Italicized text indicates descriptive language and not the actual output produced by LLMs.}
    \label{tab:full_example_2}
\end{table}

\begin{table*}[b]
\centering
% \resizebox{\linewidth}{!}{%
\small
\begin{tabular}{ll} \toprule
\textbf{Category}          & \textbf{Subcategories}                                               \\ \midrule
Drawing                    & Drawing basics, Draw pixel-perfect art, Edit paths, \dots              \\
Color                      & Adjust Colors, Select Colors, Use the Adobe Color Themes panel, \dots  \\
Painting                   & Gradients, Paint with fills and strokes, Brushes, \dots                \\
Select and arrange objects & Select objects, Move and align objects, Layers, \dots                  \\
Reshape objects            & Crop images, Transform objects, Puppet Warp, \dots                     \\
Import, export, and save   & Save artwork, Export artwork, Package files, \dots                     \\
Type                       & Create text, Using fonts in Illustrator, Format paragraphs, \dots      \\
Create special effects     & Work with effects, Graphic styles, Drop shadows, \dots                 \\
Web graphics               & Best practices, Create animations, SVG, \dots                          \\
Printing                   & Set up documents, Print with color management, Overprint, \dots        \\ \bottomrule
\end{tabular}
% }
\caption{Illustrator Functionality Overview}
\label{tab:illustrator_features}
\end{table*}

\begin{table*}[!htb]
\centering
\small
\begin{tcolorbox}[colback=gray!10, colframe=black, sharp corners=south, width=\textwidth, title=System Prompt for Task Creation in Illustrator]
% \textbf{System Prompt for Code Generation}  

You are an expert user for Adobe Illustrator. 
Your goal is to generate as many tasks as possible that are helpful and represent common needs for Illustrator users. 
\\

The generated tasks should follow the following criteria:

1. Describe these tasks so that they can be coded into a script. 

2. The tasks should not be already implemented in Illustrator.

3. The tasks should be minimally dependent on the content.
\\

Generate the tasks in plain text. Each task takes one line. Do no generate any other information such as numbering.

\end{tcolorbox}

\begin{tcolorbox}[colback=blue!5, colframe=blue!50, sharp corners=south, width=\textwidth, title=User Prompt for Task Creation with Top-down Simulation]
Give me 10 most useful tasks related to \texttt{\{subcategory\}} under the category of \texttt{\{category\}}, in Adobe Illustrator.
\\

Examples of successful tasks in the previous rounds include:

\texttt{\{a list of successful task descriptions from previous round under the same category\}}
\end{tcolorbox}

\begin{tcolorbox}[colback=blue!5, colframe=blue!50, sharp corners=south, width=\textwidth, title=User Prompt for Task Creation with Bottom-up Simulation]
Give me 10 most useful Adobe Illustrator tasks related to \texttt{\{api\}} whose description is:
\texttt{\{Names and descriptions of top k synergistic APIs\}}

Take inspiration from the following APIs and their descriptions by considering the possibility of using \texttt{\{api\}} with at least one of the following APIs.
It's okay to not use them as long as the tasks related to \texttt{\{api\}} are useful. 
\\

\texttt{\{top\_nodes\_info\}}
\\

The generated tasks should follow the following guidelines:

- Make sure the tasks are reusable.

- The tasks should be logical and reasonable to use two or more APIs together.

- The generated task should not simply be a concatenation of two API nodes, i.e., do task A and do a separate task B that doesn't closely depend on task A.

- Prioritize the usefulness of the tasks over generating exactly 10 tasks—fewer, high-quality tasks are acceptable.
\end{tcolorbox}

\caption{Prompts for Task Creation in Illustrator}
\label{prompt:creation}
\end{table*}

\begin{table*}[!htb]
\centering
\small
\begin{tcolorbox}[colback=gray!10, colframe=black, sharp corners=south, width=\textwidth, title=System Prompt for Code Generation]
% \textbf{System Prompt for Code Generation}  

You are an assistant generating ExtendScript code for Adobe Illustrator. You will be provided a query that attempts to perform an action in Illustrator. Return only the ExtendScript code snippet without additional messages, formatting, or markdown.
\\

Initialize a document to simulate this code. Generate the initialization code and task code separately in the following JSON format:
\{"init\_code": INITIALIZATION\_CODE, "code": CODE, "code\_name": "[brief description]"\}
\\

The code must follow these rules:  

1. Do not use alert. Return messages for stdout.  

2. If the task is not feasible, return \{"code": ""\}.  

3. Start error messages with "Error: ".  

4. Include necessary initialization for selecting objects.  

5. Ensure reusability of task code.  

6. Call the function you create to execute the task.  

7. Keep the initial layout minimal for clear visual results.  

8. Do not crash Illustrator.  
\end{tcolorbox}

\begin{tcolorbox}[colback=blue!5, colframe=blue!50, sharp corners=south, width=\textwidth, title=Skill Generation Prompt Structure]
Task: \texttt{\{task description\}} \\

Code from the last round:
\texttt{\{code\_last\_round\}} \\

Execution error for code from last round:
\texttt{\{error\_msg\}} \\

Visual evaluation for the outcome layout from code from the last round:
\texttt{\{validation\_last\_round\}}
\end{tcolorbox}
\caption{System Prompt for ExtendScript Code Generation in Illustrator}
\label{prompt:sys_code}
\end{table*}

\begin{table*}[!htb]
\centering
\small
\begin{tcolorbox}[colback=gray!10, colframe=black, sharp corners=south, width=\textwidth, title=System Prompt for Skill Validation]
You will be given a task description, a piece of initialization code, a layout after running the initialization code, a piece of task code, and a layout after running the task code. The context for the task is Adobe Illustrator.
\\

Your job is to judge whether the task was performed correctly or not, given the task description and the two layout figures.  
The difference between the two layout figures should reflect the result of running the task code.
\\

Sometimes, the failure reason can be that the initialization code does not generate necessary elements for the task code to run correctly, or the task code does not perform the task correctly.
\\

The output should be in JSON format:  
\{"valid": true/false, "reason": [brief reason for the judgment], "suggestion": [brief suggestion for improvement]\}
Ensure the output contains no additional messages, formatting, or markdown, so that it can be directly parsed by json.loads().
\end{tcolorbox}

\begin{tcolorbox}[colback=blue!5, colframe=blue!50, sharp corners=south, width=\textwidth, title=User Prompt for Validation with LVLM]
Task description: \texttt{\{task\_description\}}

Initialization code: \texttt{\{init\_code\}}

Task code: \texttt{\{code\}}
\\

\texttt{\{init\_layout.png\}}
\texttt{\{outcome\_layout.png\}}

\end{tcolorbox}

\caption{Prompts for Skill Validation}
\label{prompt:validation}
\end{table*}

\begin{table*}[!htb]
\centering
\small
\begin{tcolorbox}[colback=gray!10, colframe=black, sharp corners=south, width=\textwidth, title=System Prompt for Code Generation]
% \textbf{System Prompt for Code Generation}  

You are an assistant generating Office Scripts code for Microsoft Excel. Office Scripts use TypeScript to automate Excel tasks.\\

You will be provided a query that attempts to perform an action in Excel. Please return the Office Scripts code snippet without any additional message, formatting or markdown.\\

Generate the initialization code and task code separately, in the following json format: \{"init\_code": INITIALIZATION\_CODE, "code": CODE, "code\_name": "[brief description]"\}, so that it can directly be parsed by json.loads().\\

The code you write should follow the following criteria:

1. Use TypeScript/JavaScript syntax for Office Scripts.

2. Make sure to use the Excel TypeScript API (Office Scripts API) - e.g., functions like workbook.worksheets, range.format, etc.

3. When it is not possible to generate the code, return {"code": ""}. This can happen when the task is not possible to be automated or when the task is not clear.

4. The initialization code should create a clean starting environment with necessary elements to demonstrate the task code.

5. Your task code should be modular and reusable for building more complex tasks.

6. Remember to call the main function to run the task code.

7. The initial workbook layout should be simple so that the effect of running the task code is clearly visible.

8. Office Scripts for both init\_code and code typically begin with: function main(workbook: ExcelScript.Workbook) \{ ... \}
\end{tcolorbox}

\caption{System Prompt for TypeScript Code Generation in Excel}
\label{prompt:sys_code_excel}
\end{table*}

\section{More Code Generation Results}

To demonstrate our framework's adaptability, we tested \texttt{Llama-3.1-70B} and \texttt{deepseek-r1} as the code generation model. Table \ref{tab:model_strategy_success} shows the success rate on a random sample of $200$ tasks from each of the top-down and bottom-up strategies.

\end{document}